\documentclass[journal,twoside,web]{ieeecolor}
\usepackage{generic}

\usepackage{amsmath,amssymb,amsfonts}
\usepackage{algorithmic}
\usepackage{float}
\usepackage{graphicx}
\usepackage{textcomp}
\usepackage{biblatex}  
\def\BibTeX{{\rm B\kern-.05em{\sc i\kern-.025em b}\kern-.08em
    T\kern-.1667em\lower.7ex\hbox{E}\kern-.125emX}}
\begin{document}
\title{GCC-Spam: Spam Detection via GAN, Contrastive Learning, and Character Similarity Networks}
\author{Zhijie Wang, Zixin Xu, and  Zhiyuan Pan, \IEEEmembership{Member,IEEE}
\thanks{The whole repository of code, datasets and models is stored at 
\protect\url{https://github.com/Red-Scarff/GAN_spam}. }
}
\maketitle
\renewcommand{\arraystretch}{1.25}  

\begin{abstract}
The exponential growth of spam text on the Internet necessitates robust detection mechanisms to mitigate risks such as information leakage and social instability. This work addresses two principal challenges: adversarial strategies employed by spammers and the scarcity of labeled data. We propose a novel spam-text detection framework GCC-Spam, which integrates three core innovations. First, a character similarity network captures orthographic and phonetic features to counter character-obfuscation attacks and furthermore produces sentence embeddings for downstream classification. Second, contrastive learning enhances discriminability by optimizing the latent‐space distance between spam and normal texts. Third, a Generative Adversarial Network (GAN) generates realistic pseudo‐spam samples to alleviate data scarcity while improving model robustness and classification accuracy. Extensive experiments on real-world datasets demonstrate that our model outperforms baseline approaches, achieving higher detection rates with significantly fewer labeled examples.
\end{abstract}


\section{Introduction}
\label{sec:introduction}
\IEEEPARstart{T}{he} rapid growth of the Internet has been accompanied by a proliferation of spam text, which not only degrades user experience but also poses risks such as information leakage, online fraud, and social instability \cite{ref1,ref2,ref3,ref4,ref5,ref6}. Spam text detection is one of the most critical tasks in anomaly detection \cite{ref7,ref8,ref9,ref10}, and faces two primary challenges: spammers may employ adversarial behaviors, and there is a scarcity of labeled data. To address these issues, building upon existing research \cite{ref12}, this work extends and refines previous studies to introduce a novel spam detection method GCC-Spam, designed to achieve superior performance with fewer data and reduced training resources. The effectiveness of the proposed approach is demonstrated through a series of comparative experiments.

The first challenge lies in the adversarial tactics employed by spammers. Early rule-based approaches (e.g., keyword filtering) fail when spammers use semantic variants—such as rewriting the same sentence with synonyms—or character-level obfuscations, for instance replacing sensitive terms with homophones or visually similar characters to bypass censorship. Although NLP models (e.g., BERT \cite{ref11}) reduce reliance on manual keyword enumeration, their tokenization schemes depend on historical corpora and therefore cannot detect unseen adversarial patterns. Previous studies have attempted to defend against such adversarial tactics by measuring character-level similarity or computing sentence embeddings via attention mechanisms \cite{ref12}. However, these methods often overlook the distinction between normal and spam texts. To address this limitation, we introduce contrastive learning \cite{ref13}. Contrastive learning is a self-supervised approach that encourages the model to learn discriminative features in the latent space by pulling semantically similar samples closer and pushing dissimilar ones apart. By incorporating contrastive learning, our method achieves enhanced robustness against previously unseen adversarial strategies employed by spammers.

The second challenge is data scarcity. Unlike standardized tasks such as image recognition, the definition of spam varies across platforms—for instance, a message considered spam on WeChat may be deemed normal on X. This variability makes it difficult to obtain a unified, publicly annotated dataset. Furthermore, given the massive volume of daily messages and the extremely low proportion of spam, manual annotation is impractical. To address this issue, we employ Generative Adversarial Networks (GANs) to enhance labeling efficiency \cite{ref14}. GANs operate within an adversarial training framework, and in our approach, they are used to generate realistic pseudo-spam texts, augmenting the training data and improving the model’s robustness to obfuscated spam content.

To address the aforementioned challenges, this paper proposes a novel spam text detection model centered around a character similarity network. This network operates in conjunction with a character-level embedding model to transform user messages into sentence-level representations. By leveraging contrastive learning, the model enhances its ability to distinguish between spam and legitimate texts. Additionally, a GAN-based architecture is employed to generate realistic pseudo-spam samples, effectively mitigating the issue of limited labeled data while further improving classification accuracy.

In summary, the main contributions of this work are as follows:
\begin{itemize}
    \item We extend similarity-based approaches through contrastive learning, effectively countering adversarial variants of spam text.
    \item We employ a GAN-based architecture to generate pseudo-spam texts, enabling the training of a more accurate model with a smaller dataset.
    \item We conduct a comprehensive set of experiments, including ablation studies and comparisons with data augmentation, to validate the effectiveness of the proposed method.
\end{itemize}

\section{Related Work}
Spam detection research initially focused on feature engineering. Early works like \cite{ref15} utilized review centric, reviewer centric, and product centric features with Logistic Regression. Similarly, \cite{ref16} extracted diverse message features for classifiers. While enhancing feature richness, these methods struggle with adaptability, as their effectiveness hinges on predefined sets and historical patterns, often failing to capture novel adversarial actions.

The emergence of deep learning has led to more sophisticated spam detection models. For instance, hybrid approaches combining Convolutional Neural Networks (CNN) for robust feature extraction with boosting algorithms like XGBoost enhance detection against adversarial spam \cite{ref17}. These methods balance feature learning and classification accuracy, showing improved resilience to manipulated inputs. Moreover, systematic reviews emphasize advanced deep learning models like LSTMs and BERT-based architectures, which excel at capturing contextual semantic cues in phishing emails and adversarial text \cite{ref18}.

Despite these advancements, significant challenges persist in detecting evolving and novel adversarial actions. Many existing models, even deep learning ones, struggle with "concept drift" as spammers adapt tactics, making filters less potent \cite{ref20}. A key limitation is their reliance on static or outdated datasets, often lacking examples of the latest AI-generated attacks \cite{ref19}. This leads to a reactive approach, updating models only after new tactics emerge. Furthermore, while ensemble methods and adversarial training improve robustness, they primarily address known attack types or generalize within existing data, rather than proactively anticipating entirely new adversarial strategies or language shifts \cite{ref21}. The computational overhead of training and updating complex models also hinders real-time deployment and continuous adaptation to rapidly evolving spam campaigns.

\section{Proposed Method}
This section provides a detailed description of the proposed GCC-Spam method. GCC-Spam is a GAN-based spam-text detection framework that incorporates contrastive learning and a character similarity network. First, \ref{subseb:Problem Formulation} defines the spam detection problem. Next, \ref{subsec:Generator} describes the design of the GAN generator. In \ref{subsec:Discriminator}, we present the construction of the discriminator, which leverages the character similarity network. Finally, \ref{subsec:Contrastive Learning} explains how contrastive learning is integrated into the discriminator.

\subsection{Problem Formulation}
\label{subseb:Problem Formulation}

Spam text detection in Chinese presents unique challenges due to the prevalence of adversarial writing strategies specifically tailored to evade conventional lexical or rule-based filters. In this work, we define spam texts as those that aim to promote illegal content, advertisements, or harmful information (e.g., pornography, violence, political incitement), and which typically exhibit obfuscation through two key strategies:

\begin{itemize}
    \item \textbf{Character-level perturbation:} Malicious users often employ homophones or visually similar characters to substitute sensitive or banned keywords. For example, the character Wechat may be replaced with Vx , which shares a similar pronunciation, or character that is visually close. These substitutions significantly degrade the performance of traditional keyword-based filters.
    
    \item \textbf{Semantic obfuscation:} In addition to character-level changes, spam messages often retain the harmful semantic intent (e.g., explicit advertising, politically sensitive content), while avoiding the direct use of prohibited keywords. This leads to messages that are semantically offensive yet syntactically subtle.
\end{itemize}

Notably, these two forms of obfuscation—character-level substitutions and semantic-level masking—often co-occur within the same text, making detection even more challenging.

Given a corpus \( \mathcal{D} = \{(x_i, y_i)\}_{i=1}^{N} \), where \( x_i \) denotes a Chinese sentence and \( y_i \in \{0,1\} \) indicates whether \( x_i \) is spam (1) or not (0), our objective is to learn a detection function \( f_\theta: \mathcal{X} \rightarrow \{0,1\} \), parameterized by \( \theta \), that accurately classifies both clean spam and obfuscated spam variants. This function should be robust to both character-level substitutions and semantic manipulations.

\begin{figure*}[!t]
  \centering
  \includegraphics[width=\textwidth]{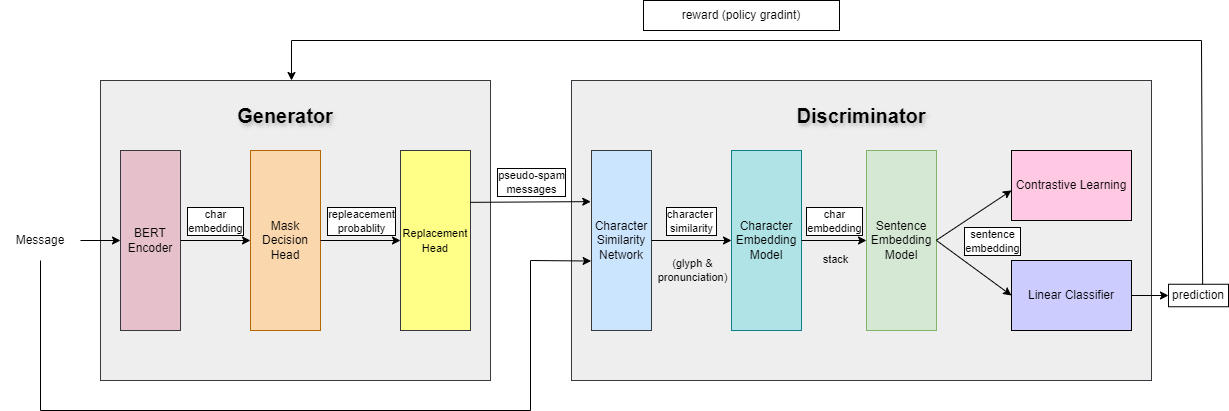}
  \caption{Flowchart of the proposed GCC-Spam method, which consists of generator and discriminator.The generator applies BERT encoder, mask decision head and replacement head. While the discriminator includes character similarity network construction module, character embedding module, sentence embedding module, contrastive learning module and spam probability prediction module.}
  \label{fig:my-wide-figure}
\end{figure*}

To address these challenges, we propose GCC-Spam, a generative adversarial framework that leverages: (i) a spam-text generator trained via policy gradient to produce deceptive perturbations based on character similarity; (ii) a discriminator enhanced by a character similarity network and (iii) a contrastive learning mechanism to improve semantic discrimination of hard negative samples. This jointly trained architecture aims to improve the robustness and generalization of spam text detection in adversarial scenarios. The flowchart is shown in \ref{fig:my-wide-figure}.

\subsection{Generator}
\label{subsec:Generator}

The generator is designed to transform spam texts into \textit{adversarial variants} that maintain semantic and visual consistency while deceiving the discriminator into predicting them as legitimate. Given an input spam sequence $s = c_1 c_2 \ldots c_n$, the generator predicts, for each position $i$, whether the character $c_i$ should be replaced and, if so, which character to substitute. The output is a modified sequence $\tilde{s} = \tilde{c}_1 \ldots \tilde{c}_n$ with potentially altered characters that evade detection.

\subsubsection{Masked Replacement Architecture}

The generator uses a BERT-based \cite{ref11} encoder to produce contextualized token representations. For each token position, two heads are applied:

\begin{enumerate}
    \item \textbf{Mask Decision Head}: A binary mask policy head predicts a probability $p_i \in [0, 1]$ of replacing the character at position $i$. A Bernoulli sample $m_i \sim \text{Bernoulli}(p_i)$ determines whether to replace $c_i$.
    
    \item \textbf{Replacement Head}: If $m_i = 1$, a character replacement distribution is generated using a Gumbel-Softmax over the vocabulary, yielding a one-hot vector indicating the replacement token $\tilde{c}_i$. If $m_i = 0$, the original character is retained.
\end{enumerate}

Formally, the final generated sequence is:

\[
\tilde{c}_i = 
\begin{cases}
\arg\max_j \text{GumbelSoftmax}(r_{ij}) & \text{if } m_i = 1 \\
c_i & \text{otherwise}
\end{cases}
\]

where $r_{ij}$ denotes the replacement logits output by the replacement head for position $i$ and vocabulary index $j$.

\subsubsection{Reinforcement Learning with Discriminator Feedback}

The generator is trained using policy gradient methods to maximize the probability of fooling the discriminator. Specifically, the reward for a generated sequence $\tilde{s}$ is defined as:

\[
R(\tilde{s}) = 1 - D(\tilde{s})
\]

where $D(\tilde{s}) \in (0, 1)$ is the discriminator’s predicted probability that $\tilde{s}$ is spam. The policy gradient loss is computed as:

\[
\mathcal{L}_{\text{RL}} = -\mathbb{E}_{\tilde{s} \sim \pi_\theta(s)} \left[ \log \pi_\theta(\tilde{s} \mid s) \cdot R(\tilde{s}) \right]
\]

In practice, the log-likelihood $\log \pi_\theta(\tilde{s} \mid s)$ is decomposed into contributions from both the mask action and the replacement token selection:

\[
\log \pi_\theta(\tilde{s}) = \sum_{i=1}^{n} \log \left( \mathbb{P}(m_i) \cdot \mathbb{P}(\tilde{c}_i \mid m_i) \right)
\]

where $\mathbb{P}(m_i)$ is the Bernoulli probability of masking, and $\mathbb{P}(\tilde{c}_i \mid m_i)$ is either the replacement probability from Gumbel-Softmax or a delta function for identity if no replacement.

\subsubsection{Similarity Loss Regularization}

To ensure character substitutions remain plausible, the generator incorporates a \textbf{character-level similarity loss} using the character similarity matrix shared with the discriminator. For each replaced position, a similarity score between the original and the selected substitute is computed. The similarity loss encourages the model to select visually and phonetically similar substitutes:

\[
\mathcal{L}_{\text{sim}} = \frac{1}{|V|} \sum_{i \in V} \left(1 - \text{sim}(c_i, \tilde{c}_i)\right)
\]

where $V$ is the set of positions where $c_i$ is a Chinese character and a replacement was considered. The final training loss is a weighted sum of the reinforcement learning loss and the similarity loss:

\[
\mathcal{L}_{\text{G}} = \mathcal{L}_{\text{RL}} + \lambda_{\text{sim}} \cdot \mathcal{L}_{\text{sim}}
\]

\subsubsection{Summary}

This design enables the generator to produce contextually coherent and semantically consistent adversarial spam texts. By leveraging BERT-based representations \cite{ref11}, hard Gumbel-Softmax sampling, character similarity constraints, and discriminator feedback via policy gradients, the generator learns to craft effective perturbations that evade spam detection without disrupting text readability.

\subsection{Discriminator}
\label{subsec:Discriminator}

The discriminator in our proposed framework adopts the architecture of the LZASD method, as introduced in \cite{ref12}, specifically designed to tackle spam text detection by addressing adversarial deformations employed by spammers. This approach leverages character-level features to capture subtle patterns, such as deformed characters, that traditional word-based models often miss. The discriminator processes a user message denoted as 
\( c = "c_1 c_2 \ldots c_n" \)
, where \( c_i \) represents the \( i \)-th character, aiming to predict the probability \( p_c \) that the message is spam. Below, we detail the core components, their principles, and mathematical formulations, highlighting how the discriminator effectively handles spam detection challenges.

\subsubsection{Character Similarity Network}

The cornerstone of the discriminator is the character similarity network, constructed over a Chinese character repository \( C = \{c_1, c_2, \ldots, c_m\} \), where \( c_i \) denotes the \( i \)-th character in the dictionary. This network quantifies similarities between characters based on two primary dimensions: glyph (e.g., stroke structure, such as left-right or up-down configurations) and pronunciation (e.g., pinyin components like initials, finals, and tones). Spammers often replace characters with visually or phonetically similar ones to evade detection, and this network aims to capture such adversarial substitutions. The overall similarity between two characters \( c_i \) and \( c_j \) is defined as
\[ \text{sim}(c_i, c_j) = \max(\text{sim}_g(c_i, c_j), \text{sim}_p(c_i, c_j)) \]
where \( \text{sim}_g(c_i, c_j) \in [0, 1] \) measures glyph similarity, and \( \text{sim}_p(c_i, c_j) \in [0, 1] \) evaluates pronunciation similarity. For glyph similarity, a 7-bit glyph code encodes structural features: the first bit captures the basic structure (e.g., left-right, up-down), bits 2 to 6 represent the four-corners code, and the last bit denotes stroke count. Pronunciation similarity uses a 4-bit code, encoding initials, finals, and tones. By taking the maximum of these two metrics, the model accounts for deformations that are either visually or phonetically similar, a critical feature for detecting disguised spam.

\subsubsection{Character Embedding}

To address the limitation of traditional word embeddings, which fail to recognize novel deformed words, the discriminator generates character embeddings that incorporate semantic and emotional features. For a character \( c_i \), the embedding is computed as a weighted sum of embeddings of similar characters within the similarity network: 
\[ \text{emb}(c_i) = \frac{\sum_{c_k \in N(c_i)} \text{freq}(c_k) \cdot \text{w2v}(c_k)}{\sum_{c_k \in N(c_i)} \text{freq}(c_k)} \]
 Here, \( N(c_i) = \{c_k | c_k \in C, \text{sim}(c_i, c_k) > \rho\} \) represents the set of characters similar to \( c_i \), with similarity exceeding a threshold \( \rho \in [0, 1] \). The term \( \text{freq}(c_k) \in [0, 1) \) denotes the frequency of character \( c_k \) in the repository, reflecting its prevalence, while \( \text{w2v}(c_k) \in \mathbb{R}^d \) is the embedding vector learned via the Word2Vec model, capturing contextual semantic information. This weighted sum approach ensures that the embedding of \( c_i \) integrates features from a cluster of similar characters, making the representation robust to deformations—whether the deformed character exists in the training data or not. The use of Word2Vec is justified by its efficiency and stability in modeling semantic relationships based on character context.

\subsubsection{Sentence Embedding}

With character embeddings generated, the model constructs a sentence embedding to represent the entire message \( s = "c_1 c_2 \ldots c_n" \). A self-attention layer is employed to capture relationships between characters, enhancing the model’s ability to identify spam-related patterns. The process begins by computing attention scores: 
\[ \hat{\alpha}_{i,j} = \frac{\text{emb}(c_i) \otimes \text{emb}(c_j)}{\sqrt{d}} \]
where \( \text{emb}(c_i) \) and \( \text{emb}(c_j) \) are embeddings of characters \( c_i \) and \( c_j \), and \( d \) is the embedding dimension for scaling. These scores are normalized using a softmax function: 
\[ \alpha_{i,j} = \frac{e^{\hat{\alpha}_{i,j}}}{\sum_{k=1}^n e^{\hat{\alpha}_{i,k}}} \]
producing attention weights that reflect the importance of each character relative to others. The weighted sum of embeddings is then computed as
\[ m_i = \sum_{j=1}^n \hat{\alpha}_{i,j} * \text{emb}(c_j) \]
capturing contextual interactions. Finally, the sentence embedding is obtained by averaging:
\[ \text{emb}(s) = \frac{1}{|s|} \sum_{c_i \in s} m_i \]
This representation encodes the semantic and thematic content of the message, providing a robust input for classification.

\subsubsection{Classifier}

The sentence embedding \( \text{emb}(s) \) is subsequently fed into a binary classifier, such as logistic regression or a gradient boosting decision tree, to predict the spam probability: 
\[ p = \text{classifier}(\text{emb}(s)) \]
where \( p \in (0, 1) \) indicates the likelihood of the message being spam. The classifier is trained using a log-loss function to optimize prediction accuracy. This loss function penalizes deviations from the ground truth, ensuring the model learns to distinguish spam from legitimate messages effectively.

\subsection{Contrastive Learning}
\label{subsec:Contrastive Learning}

To further enhance the discriminative power of the discriminator, we incorporate supervised contrastive learning (InfoNCE \cite{ref23}) into the embedding space. Let 
\[
\mathcal{B} = \{(s_i, y_i)\}_{i=1}^{N}
\]
denote a mini-batch of $N$ samples, where each $s_i$ is a sentence (either spam or legitimate) and $y_i \in \{0,1\}$ is its label ($1$ for spam, $0$ for legitimate). After obtaining the sentence embedding
\[
\mathbf{z}_i = \mathrm{emb}(s_i) \in \mathbb{R}^d,
\]
we define the cosine similarity between two embeddings $\mathbf{z}_i$ and $\mathbf{z}_j$ as
\[
\mathrm{sim}(\mathbf{z}_i, \mathbf{z}_j) \;=\; \frac{\mathbf{z}_i^\top \mathbf{z}_j}{\|\mathbf{z}_i\| \,\|\mathbf{z}_j\|}.
\]
For each anchor index $i$, let 
\[
\mathcal{P}(i) = \{\,p \mid y_p = y_i,\;p \neq i\}
\]
be the set of indices in $\mathcal{B}$ sharing the same class label as $s_i$. Then, the InfoNCE loss for the entire batch is
\[
\mathcal{L}_{\mathrm{CL}}
=
- \frac{1}{N} \sum_{i=1}^{N} 
  \frac{1}{|\mathcal{P}(i)|} 
  \sum_{p \in \mathcal{P}(i)} 
    \log \frac{\exp\bigl(\mathrm{sim}(\mathbf{z}_i,\mathbf{z}_p)/\tau\bigr)}
                 {\sum\limits_{a=1,\,a\neq i}^{N} 
                  \exp\bigl(\mathrm{sim}(\mathbf{z}_i,\mathbf{z}_a)/\tau\bigr)}\,
                  \],
where $\tau > 0$ is a temperature hyperparameter. This loss encourages embeddings of samples from the same class (either spam or normal) to be pulled closer in the latent space, while pushing embeddings of different classes farther apart. 

In practice, during each training iteration, we jointly minimize the classification loss and the contrastive loss:
\[
\mathcal{L} = \mathcal{L}_{\text{CE}} + \lambda_{\mathrm{CL}}\;\mathcal{L}_{\mathrm{CL}},
\]
where $\lambda_{\mathrm{CL}}$ balances the two objectives. By integrating contrastive learning into the discriminator, GCC-Spam achieves a more pronounced separation between spam and normal text representations, thereby improving robustness against adversarially deformed inputs.

\section{Experiments}
\label{sec:experiments}

This section presents and analyzes the experimental results of the proposed GCC-Spam model. We compare our approach against three baselines: a LZASD architecture discriminator without GAN or contrastive learning (Baseline), a spam detector enhanced by GAN-based adversarial training, and two data augmentation strategies (static and dynamic, which will be introduced in Appendix~\ref{appendix:augmentation}). All models are evaluated on a Chinese spam text dataset containing character-level and semantic-level perturbations. Metrics include precision, recall, F1-score, and overall accuracy.

\subsection{Performance Comparison}

Table~\ref{tab:baseline} shows that the baseline model achieves an overall accuracy of 94.0\%, with a macro F1-score of 0.93. Although this result is reasonably strong, it reveals limitations in handling adversarially perturbed spam, particularly in the recall of non-spam (class 0), which remains at 0.87.

\begin{table}[H]
\centering
\caption{Performance of Baseline Discriminator}
\label{tab:baseline}
\begin{tabular}{|l|c|c|c|c|}
\hline
Class / Metric & Precision & Recall & F1-score & Support \\
\hline
0 & 0.93 & 0.87 & 0.90 & 2498 \\
1 & 0.94 & 0.97 & 0.96 & 5506 \\
\hline
Accuracy & & & 0.94 & 8004 \\
Macro Avg & 0.94 & 0.92 & 0.93 & 8004 \\
Weighted Avg & 0.94 & 0.94 & 0.94 & 8004 \\
\hline
\end{tabular}
\end{table}

In contrast, Table~\ref{tab:gan} illustrates that our GCC-Spam model significantly improves detection performance. It achieves an accuracy of 97.52\%, a macro F1-score of 0.9712, and class-specific F1-scores of 0.9605 (non-spam) and 0.9819 (spam), respectively. These results indicate that GCC-Spam not only enhances the model’s capability to detect spam with high precision but also reduces false positives in benign samples, benefiting from the integration of character similarity modeling and contrastive learning.

\begin{table}[H]
\centering
\caption{Performance of GCC-Spam (5 epoch)}
\label{tab:gan}
\begin{tabular}{|l|c|c|c|c|}
\hline
Class / Metric & Precision & Recall & F1-score & Support \\
\hline
0 & 0.9566 & 0.9644 & 0.9605 & 5000 \\
1 & 0.9838 & 0.9801 & 0.9819 & 11007 \\
\hline
Accuracy & & & 0.9752 & 16007 \\
Macro Avg & 0.9702 & 0.9723 & 0.9712 & 16007 \\
Weighted Avg & 0.9753 & 0.9752 & 0.9752 & 16007 \\
\hline
\end{tabular}
\end{table}

Compared to both static and dynamic data augmentation methods, GCC-Spam demonstrates superior generalization. The dynamic augmentation approach (Table~\ref{tab:dynamic}) improves upon the baseline with 97.21\% accuracy, but still underperforms GCC-Spam. Static augmentation (Table~\ref{tab:static}), despite leveraging a larger dataset (52,935 samples), achieves only 88.93\% accuracy and suffers from significantly lower F1-scores, highlighting the inefficiency of naive data expansion without task-aware modeling.

\begin{table}[H]
\centering
\caption{Performance of Dynamic Data Augmentation (10 epoch)}
\label{tab:dynamic}
\begin{tabular}{|l|c|c|c|c|}
\hline
Class / Metric & Precision & Recall & F1-score & Support \\
\hline
0 & 0.9423 & 0.9702 & 0.9561 & 5000 \\
1 & 0.9863 & 0.9730 & 0.9796 & 11007 \\
\hline
Accuracy & & & 0.9721 & 16007 \\
Macro Avg & 0.9643 & 0.9716 & 0.9678 & 16007 \\
Weighted Avg & 0.9725 & 0.9721 & 0.9722 & 16007 \\
\hline
\end{tabular}
\end{table}

\begin{table}[H]
\centering
\caption{Performance of Static Data Augmentation (10 epoch)}
\label{tab:static}
\begin{tabular}{|l|c|c|c|c|}
\hline
Class / Metric & Precision & Recall & F1-score & Support \\
\hline
0 & 0.8632 & 0.8543 & 0.8588 & 20848 \\
1 & 0.9060 & 0.9121 & 0.9090 & 32087 \\
\hline
Accuracy & & & 0.8893 & 52935 \\
Macro Avg & 0.8846 & 0.8832 & 0.8839 & 52935 \\
Weighted Avg & 0.8891 & 0.8893 & 0.8892 & 52935 \\
\hline
\end{tabular}
\end{table}

\subsection{Efficiency and Deployment Considerations}

In addition to performance gains, GCC-Spam offers several practical advantages:

\begin{itemize}
    \item \textbf{Cost-efficiency:} Unlike data augmentation strategies or large-scale pretrained detectors that require expensive API calls or large computing resources, GCC-Spam can be trained end-to-end on modest hardware and reused across multiple detection tasks without additional labeling effort.
    \item \textbf{Faster convergence:} Compared to static and dynamic data augmentation methods, GCC-Spam converges more quickly and requires fewer labeled samples to achieve comparable results, effectively reducing training time and data dependency.
    \item \textbf{Lightweight design:} The architecture of GCC-Spam contains significantly fewer parameters than existing large models, making it easy to deploy in real-time systems or on edge devices where resources are limited.
\end{itemize}

\subsection{Summary of Experimental Insights}

The experimental results validate the effectiveness and practicality of the proposed GCC-Spam framework. Key observations include:

\begin{itemize}
    \item GCC-Spam consistently outperforms the baseline and both static and dynamic data augmentation approaches across most evaluation metrics.
    \item The model excels in detecting both character-level perturbations and semantically misleading spam through the synergy of contrastive learning and a character similarity network.
    \item It achieves state-of-the-art performance while maintaining low computational and deployment costs.
    \item Its fast convergence and low parameter count make it a scalable and sustainable solution for long-term spam detection in dynamic environments.
\end{itemize}

\section{Conclusion}

In this work, we have presented GCC-Spam, a novel spam-text detection framework that applies GAN architecture and integrates contrastive learning and a character similarity network to address two fundamental challenges: adversarial evasion by spammers and the scarcity of labeled data.

Empirical evaluation demonstrates that GCC-Spam consistently outperforms baseline methods. Notably, GCC-Spam achieves higher detection accuracy with significantly fewer labeled examples and converges more rapidly during training. In addition, the lightweight architecture facilitates deployment in resource‐constrained environments without sacrificing detection performance. These characteristics make GCC-Spam particularly well suited for long‐term, real‐time spam monitoring where both computational efficiency and adaptability to evolving adversarial strategies are critical.

Looking forward, several avenues remain for further research. For instance, extending the contrastive learning paradigm to a semi‐supervised setting could leverage large volumes of unlabeled text more effectively. In addition, exploring cross‐domain generalization—such as adapting GCC-Spam to multilingual or multimodal (e.g., image‐embedded) spam—will further validate the broad applicability of our approach. We anticipate that the modular design of GCC-Spam will facilitate these extensions, paving the way for increasingly robust and efficient spam detection systems.

\appendices

\section{Enhanced Discriminator with Data Augmentation}
\label{appendix:augmentation}

The LZASD architecture discriminator, as described earlier, effectively captures adversarial deformations in spam text detection. However, its performance can be limited by the scarcity and lack of diversity in training data, particularly when facing novel or varied spam patterns. To address this, we propose an enhanced version of the discriminator by incorporating data augmentation, improving robustness and generalization. This appendix details the data augmentation approach, its principles, benefits, and integration into the training process.

The core idea of data augmentation in this context is to expand the training dataset by generating synthetic samples that mimic the characteristics of spam and normal texts, especially those misclassified by the model. This approach leverages the observation that spammers often employ adversarial tactics, such as replacing characters with similar glyph or pronunciation, to evade detection. By analyzing prediction errors and generating similar samples, the model can learn from a broader and more diverse set of examples, enhancing its ability to handle unseen variations. 

Prediction errors are analyzed by evaluating the model on the train set, identifying misclassified samples (i.e., texts where the predicted label differs from the true label, “spam” or “normal”). These error samples, along with their true labels and prediction probabilities, are saved for further processing. Data augmentation then employs a large language model (LLM) API, such as DeepSeek-R1-0528. The augmentation process generates new samples by feeding a subset of error samples (e.g., 100 samples) to the LLM, with a prompt instructing it to create similar texts. The prompt specifies requirements: (1) generated texts should match the style and features of examples, (2) semantics can be close but not identical, (3) character substitutions based on glyph or pronunciation similarity are encouraged, (4) text length should be similar, (5) spam texts may be unnatural to mimic evasion tactics, while normal texts should be natural, and (6) the number of generated samples (e.g., 500, with 250 spam and 250 normal) is specified.

The training process integrates augmented data as follows. In the dynamic augmentation process, for each epoch, if an augmented data file exists, it is read and parsed into spam and normal samples. These augmented samples are accumulated across epochs to enrich the dataset. In each batch, a portion of samples is randomly drawn from augmented data, prioritizing these to introduce diversity, while the remainder comes from original training data, maintaining a balance between spam and normal texts. The batch is further refined to ensure balanced representation of classes using a pairing mechanism. Sentence embeddings are generated via self-attention, and the model computes a combined loss. After each epoch, new augmented data is generated via the LLM, saved, and incorporated into subsequent training.

In addition, static data augmentation is employed before training starts, with LLM API called to generate a large number of synthetic samples that mimic both spam and normal texts. These augmented samples are incorporated directly into the training dataset, with no distinction made between the original and augmented samples. This method effectively expands the training dataset without needing to differentiate between the two types of samples, allowing the model to train on a more diverse set of data from the beginning.

The benefits of data augmentation approach are multifaceted. First, it significantly expands the training dataset, introducing diverse samples that simulate adversarial tactics, such as character substitutions, thereby improving the model’s ability to generalize to unseen spam patterns. Second, by focusing on misclassified samples, it targets the model’s weaknesses, enhancing robustness against challenging cases. Third, the integration of augmented data allows continuous learning from varied examples, reducing overfitting to the original dataset. However, challenges include dependency on the quality of the LLM-generated samples, potential introduction of noise, and increased computational cost due to API calls. This enhanced discriminator leverages data augmentation to achieve superior spam detection performance, forming a comparison with our GCC-Spam method, and the experimental results are in the Section~\ref{sec:experiments}.

\printbibliography

\end{document}